\documentclass{article} 
\usepackage{iclr2015,times}
\usepackage{hyperref}
\usepackage{url}

\usepackage{amsmath}
\usepackage[pdftex]{graphicx}
\usepackage{bm}

\newtheorem{algorithm}[equation]{Algorithm}
\newcommand{\bb}[1]{\bm{\mathrm{#1}}}
\newcommand{\mat}[1]{\ensuremath{\mathbf{#1}}}
\def\Tr{\mathrm{T}}

\title{Improving approximate RPCA with a k-sparsity prior}

\author{
  Maximilian Karl \& Christian Osendorfer \\
  Fakult\"at f\"ur Informatik\\
  Technische Universit\"at M\"unchen\\
  \texttt{\{karlma,osendorf\}@in.tum.de}
}
\begin{document}

\maketitle
\maketitle

\begin{abstract}
    A process centric view of robust PCA (RPCA) allows its fast approximate 
    implementation based on a special form of a deep neural network with weights shared
    across all layers. However, empirically this fast approximation to RPCA fails to find
    representations that are parsemonious. We resolve these bad local minima by relaxing
    the elementwise L1 and L2 priors and instead utilize a structure inducing k-sparsity prior.
    In a discriminative classification task the newly learned representations outperform these
    from the original approximate RPCA formulation significantly.
\end{abstract}

\section{Introduction}

In this work an efficient implementation of sparse coding is
evaluated. Sparse coding is an optimisation problem, where
it is to find a sparse representation of given data. Not only
is it diffcult to find a mapping from the sparse code to
the data but the search for the perfect sparse code
for a single datapoint is a single extra optimisation procedure.
This process is very time intensive because you need to
optimize two problems one after another. The idea behind
an efficient implementation of this sparse coding problem
comes from \cite{sprechmann_learning_2012} \cite{gregor_learning}. A gradient descent algorithm
optimizing the sparse code with many iteration is transformed
into a neural network with very few layers, each representing 
one iteration of the gradient descent algorithm. This network is then
trained using the same objective function as used for creating the
gradient descent iterates creating an efficient version of the initial
optimisation procedure.

The Robust Principal Component Analysis (RPCA or Robust PCA)\cite{sprechmann_learning_2012} \cite{candes_robust_2011}
version of such an efficient sparse coding network is evaluated.
An own optimised version of this algorithm producing much sparser latent codes is
presented and evaluated.

\section{Robust PCA}

The motivation behind Robust PCA is the decomposition of
a large matrix into a low rank matrix and a sparse matrix \cite{candes_robust_2011}. The sparse matrix is also
called outlier matrix, therefore also the name Robust PCA. This decomposition can
be formulated as follows:

\begin{equation}
    M = L_0 + S_0
\end{equation}

where $M$ is the large data matrix, $L_0$ is the low rank matrix and $S_0$ is the sparse
outlier matrix.

In the often used Principal Component Analysis (PCA) a similar problem is solved. However
normal PCA features no outlier matrix. So it tries to minimize $\| M-L \|_F$ subject to $rank(L) \le k$
where only small disturbances are allowed. Single large disturbances could render the low rank matrix
different from the true low rank matrix. Through introducing an outlier matrix these corruptions could
be eliminated helping the low rank matrix capture the information of the real data.

\section{Efficient Sparse Coding}

The efficient RPCA algorithm\cite{sprechmann_learning_2012} uses the same objective function as
the original RPCA formulation\cite{candes_robust_2011}:

\[
\frac{1}{2} \|X-DS-O\|_F^2 + \frac{\lambda_{\ast}}{2}( \|D\|_F^2 + \|S\|_F^2) + \lambda\|O\|_1
\]

This objective function is transformed into a neural network by first deriving the proximal descent iterations.
This means computing the gradient of the smooth part of the objective function wrt $S$ and $O$ and
computing a proximal operator out of the non-smooth part. The proximal operator is defined as followed \cite{sprechmann_learning_2012} \cite{bach_convex}:

\begin{equation}
\pi_{\frac{\lambda}{\alpha} \psi}(z) = \underset{u \in R^m}{argmin} \frac{1}{2} \| u - z \|_2^2 + \frac{\lambda}{\alpha} \psi (u)
\end{equation}

where $\psi (u)$ is the non-smooth part of the objective function.

Constructing the proximal descent of this objective function results in the following algorithm:

\newcommand{\For}[2]{for {#1} do {#2}}

\begin{algorithm}{RPCA Proximal Descent. With $\pi_{\bb{t}}(\bb{b}) = sign(\bb{b})max(0,|\bb{b}|-\bb{t})  $. Taken from \cite{sprechmann_learning_2012}. $x$ is the input, $D$ is the dictionary, $l$ and $o$ are the low-rank approximation and the outlier \label{alg:rpca}}

Define $\bb{H} = \bb{I} - \frac{1}{\alpha}\left(
                                            \begin{array}{cc}
                                              \mat{D}_0^\Tr \mat{D}_0 + \lambda_{\ast}I &  \mat{D}_0^\Tr \\
                                              \mat{D}_0 & \bb{I} \\
                                            \end{array}
                                          \right)$, \\
$\bb{W} = \frac{1}{\alpha}\left(
                                            \begin{array}{c}
                                              \mat{D}_0^\Tr\\
                                              \bb{I} \\
                                            \end{array}
                                          \right)$, and $\bb{t} = \frac{\lambda}{\alpha} \left(
                                            \begin{array}{c} 0 \\ 1 \\ \end{array}
                                          \right)$.

Initialize $\mat{z}^0 = \bb{0}$, $\bb{b}^0 = \bb{W} \bb{x}$.

\For{$k=1,2,\dots$ until convergence}
{

$\begin{array}{ll} \vline & \begin{array}{l}
  \bb{z}^{k+1} = \pi_{\bb{t}}( \bb{b}^k ) \\
  \bb{b}^{k+1} = \bb{b}^k + \bb{H} (\bb{z}^{k+1} - \bb{z}^k)
  \end{array}\end{array}$

}

Split $\bb{z}^{k+1} = (\bb{s}; \bb{o})$ and output $\bb{l} = \mat{D}_0\bb{s}$.
\end{algorithm}

Because this iterative algorithm is costly we need an efficient implementation
of it. This is done by unrolling the loop and building a neural network with
a fixed size out of it\cite{sprechmann_learning_2012} \cite{gregor_learning}. Each layer of the neural network represents one
iteration of the proximal splitting algorithm. The matrices $H$ and $W$
can be interpreted as weight matrices.
The parameters $W$, $H$, $t$ and $D$ can now be trained using
standard optimization techniques from neural networks. This fine-tuning
creates iterations that are more efficient than the original proximal splitting method.
One could either train all parameter at once or constrain $H$, $W$ and $t$ to train only the dictionary $D$.
Another possibility is to train different $H$ and $W$ for every layer
to create a more powerful model. The focus of this work was put
on first training the dictionary and fixing all other parameters to the
proximal splitting algorithm initialisation.

\section{Instabilities}

During evaluating this efficient RPCA network on MNIST some problems arised. The objective function
consists of a reconstruction term, a sparsity term and an outlier term. Optimizing
this RPCA Network resulted in a decrease of this objective function. At the same time
the reconstruction error in this objective function increased which implies a bad
reconstruction from the sparse code. However the output of the
network featured every detail of the desired output. This came from the fact that the
network saved all information in the outlier matrix. The sparse code was therefore completely
blank. Changing the parameters to stabilize this problem resulted in non-sparse codes and
good reconstruction which is also not desired.

The problem lies in the regularizer for the sparse code. Here for the sparse code the l2-norm
and for the sparse outlier the l1-norm was used. Both of them only act on single elements
of the sparse code and outlier. A regularizer selecting some of these elements and applying
a regular l1-norm to only them would solve this problem.

\section{k-Sparse Regularizer}

The solution to this problem is to use the k-sparse function from k-sparse autoencoders\cite{makhzani_k-sparse_2013}
and taking it as a base for the new regularizer. The k-sparse function selects the k-largest elements of
an array and sets all other elements to zero. This makes it an ideal candidate for building a
regularizer which only applies a l1-norm to some of these elements. The new norm is defined
as follows:

\[
\|S-kSparse(S,k_{\ast})\|_1
\]

It is a l1-norm between the k-sparse operator applied to the sparse code and the
sparse code itself. $S$ is the sparse code and $k_{\ast}$ the parameter
regulating the number of non-zero elements. This regularizer now protects
all k-largest elements of the sparse code from the l1-norm.

\section{Efficient k-Sparse Coding}

Instead of just applying the l1-norm to the outliers we now use the k-sparse norm.
This allows a fixed amount of information to be stored in $O$. This amount of
information can be controlled by the parameter $k$. This prevents the network from
stroring all information in the outlier matrix and leaving the sparse code
empty. To further improve the sparsiness of the sparse code the norm
was also applied to $S$. Using this k-sparse prior the overall sparse
coding objective function changes to:

\[
\frac{1}{2} \|X-DS-O\|_F^2 + \lambda_{\ast} \|S-kSparse(S,k_{\ast})\|_1 + \lambda\|O-kSparse(O,k)\|_1
\]

Of course the optimal parameter from RPCA may not be the perfect parameters for the k-sparse
instance but it has shown that this new setting is much more robust against
variations in the parameters. Also $\|D\|_F^2$ is not present anymore because
its minimisation was entangled with the minimisation of $\|S\|_F^2$ since
they represent together the minimisation of the rank of $DS$ \cite{sprechmann_learning_2012}.

When this k-sparse prior is used in the objective function
and processed using the proximal descent framework something interesting
happens. Instead of just applying the shrinkage function to every element now the k-largest values are protected from the shrinkage function. Instead of applying the k-sparse operator directly on the
sparse code as in the k-Sparse Autoencoder setting \cite{makhzani_k-sparse_2013}
here the k-sparse function is applied as some kind of soft manner.

The derivation of the proximal operator for the k-sparse
coding case with $\psi(u) = \lambda_{\ast} \|S-kSparse(S,k_{\ast})\|_1 + \lambda\|O-kSparse(O,k)\|_1$
can be splitted in two separate proximal operators since $S$ and $O$ are independent
parts of the vector $u$. The derivation of the proximal operator for
one single of these vector parts:

\begin{eqnarray*}
\pi_{\frac{1}{\alpha} \psi}(z) &=& \underset{u \in R^m}{argmin} \frac{1}{2} \| u - z \|_2^2 + \frac{1}{\alpha} \psi (u)\\
                             0 &=& \nabla_u (\frac{1}{2} \| u - z \|_2^2  + \frac{\lambda}{\alpha} | u - kSparse(u,k) |_1)\\
                             0 &=& u - z + \nabla_u \frac{\lambda}{\alpha} | u - kSparse(u,k) |_1\\
                             z &=& u + \frac{\lambda}{\alpha} sgn(u - kSparse(u,k)) \\
\end{eqnarray*}

This function needs to be inverted. For elements for which $u = kSparse(u,k)$ applies
the proximal operator is the identity function. In the other case $kSparse(u,k) = 0$ the
proximal function is the same as in the original RPCA case. This soft k-sparse shrinkage
function derived from the objective function looks like this:

\[
\kappa_{\bb{t}, \bb{k}}(\bb{b}) = \tau_{\bb{t}, \bb{k}}(\bb{b}) - \tau_{\bb{t}}(kSparse(\bb{b},\bb{k})) + kSparse(\bb{b},\bb{k})
\]

where $\tau_{\bb{t}, \bb{k}}(\bb{b})$ is the original soft shrinkage function applied at every
iteration. $kSparse(\bb{b},\bb{k})$ is the original k-sparse function from \cite{makhzani_k-sparse_2013}.
The complete algorithm looks very similiar to the RPCA case:

\begin{algorithm}{k-Sparse Proximal Descent. With $\pi_{\bb{t}, \bb{k}_{\ast}, \bb{k}}(\bb{b}) = \left(
                                            \begin{array}{c} \kappa_{\bb{t}, \bb{k}_{\ast}}(\bb{S}) \\ \kappa_{\bb{t}, \bb{k}}(\bb{O})\\ \end{array}
                                          \right)  $. Taken from \cite{sprechmann_learning_2012} and modified to match the k-Sparse Proximal Descent. \label{alg:ksrpca}}

Define $\bb{H} = \bb{I} - \frac{1}{\alpha}\left(
                                            \begin{array}{cc}
                                              \mat{D}_0^\Tr \mat{D}_0 &  \mat{D}_0^\Tr \\
                                              \mat{D}_0 & \bb{I} \\
                                            \end{array}
                                          \right)$, \\
$\bb{W} = \frac{1}{\alpha}\left(
                                            \begin{array}{c}
                                              \mat{D}_0^\Tr\\
                                              \bb{I} \\
                                            \end{array}
                                          \right)$, and $\bb{t} = \frac{1}{\alpha} \left(
                                            \begin{array}{c} \lambda_{\ast} \\ \lambda \\ \end{array}
                                          \right)$.

Initialize $\mat{z}^0 = \bb{0}$, $\bb{b}^0 = \bb{W} \bb{x}$.

\For{$k=1,2,\dots$ until convergence}
{

$\begin{array}{ll} \vline & \begin{array}{l}
  \bb{z}^{k+1} = \pi_{\bb{t}, \bb{k}_{\ast}, \bb{k}}( \bb{b}^k ) \\
  \bb{b}^{k+1} = \bb{b}^k + \bb{H} (\bb{z}^{k+1} - \bb{z}^k)
  \end{array}\end{array}$

}

Split $\bb{z}^{k+1} = (\bb{s}; \bb{o})$ and output $\bb{l} = \mat{D}_0\bb{s}$.
\end{algorithm}

The differences to the RPCA algorithm is not only the change in the activation function
but also in the matrix $H$. This matrix does not include the parameter $\lambda_{\ast}$
anymore. This comes from the fact that the norm of the sparse code is not part
of the smooth part of the objective function but now of the non-smooth function
which only affects the proximal operator. Instead $t$ now incorperates $\lambda_{\ast}$
since $t$ is derived from the non-smooth part of the objective function.

\section{Experiments}
\label{chapter:exp}

For the experiments the MNIST\footnote{yann.lecun.com/exdb/mnist/} Dataset was used.
The dataset contained no outliers. We were only interested in the relative
performance of the two algorithms. The efficient RPCA
and efficient k-sparse coding model were both trained unsupervised on this dataset.
To be able to compare the quality of these different representations the classification
error was choosen. For each representation a supervised logistic regressor was trained
to classify the correct type of digit. The errors for this experiment are shown in \ref{tab:regression}.
They represent the number of falsely classified digits in percent.
The k-sparse coding model is producing suprisingly lower errors compared to RPCA.
This shows k-sparse is producing hidden representations better suited for classification
using linear classifiers. The change of the $k$ parameter shows small changes in
classification error and allows some fine-tuning. Very small values of $k$ results
in high error rates since less information can be stored in this small number of
non-zero hidden values. Whereas very high values would result in similar error rates
to the RPCA case since then the objective function is then more similar to the one of
k-sparse coding.

\begin{figure*}[]
    \centering
    \includegraphics[width=\textwidth]{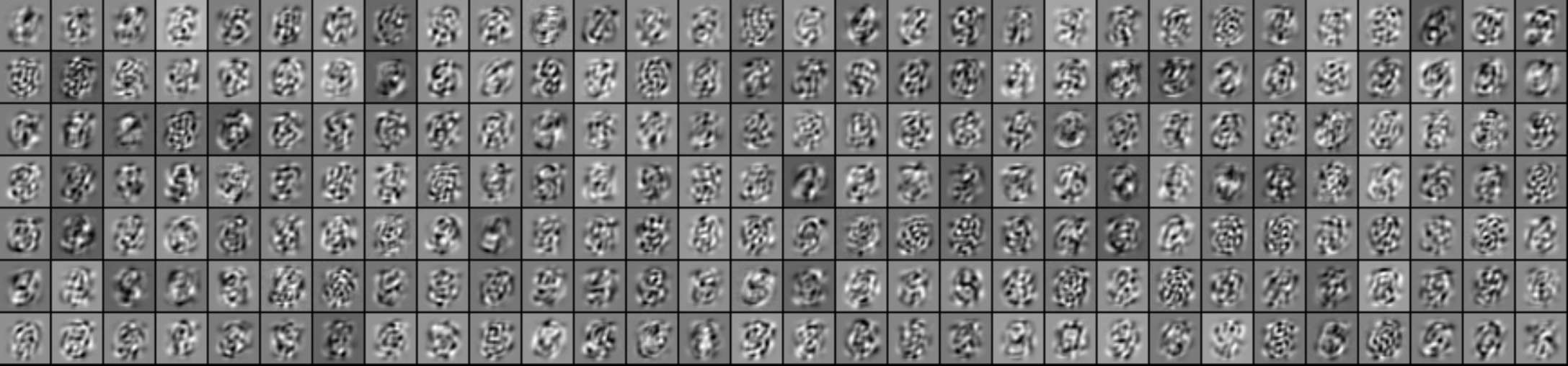}
      \caption{Some entries from the dictionary for RPCA with 1000 hidden units}
  \label{fig:dict_rpca}
\end{figure*}

\begin{figure*}[]
    \centering
    \includegraphics[width=\textwidth]{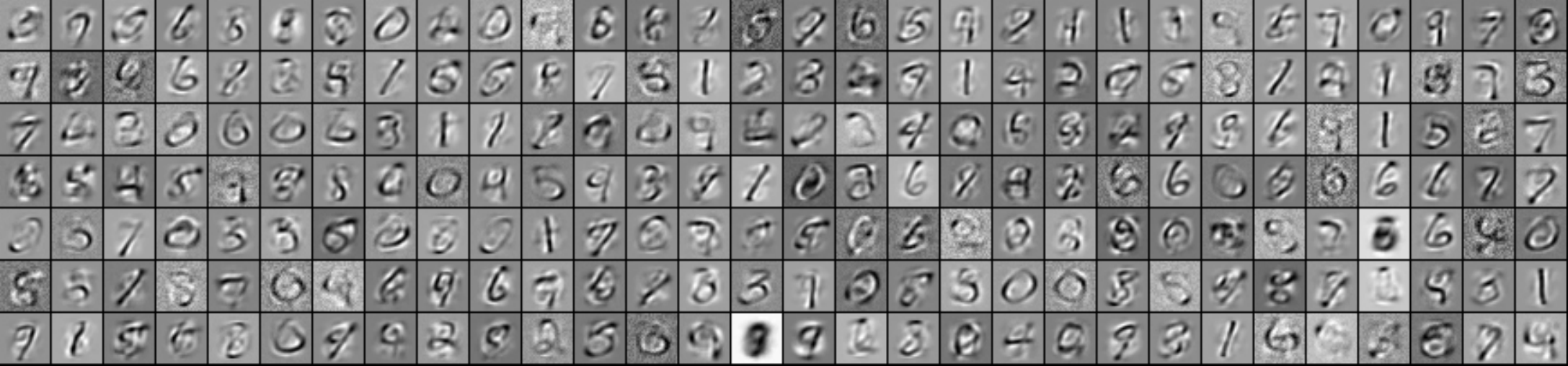}
      \caption{Some entries from the dictionary using the new k-sparse prior with 1000 hidden units and $k=20$}
  \label{fig:k_20_rpca}
\end{figure*}

\begin{figure*}[]
    \centering
    \includegraphics[width=\textwidth]{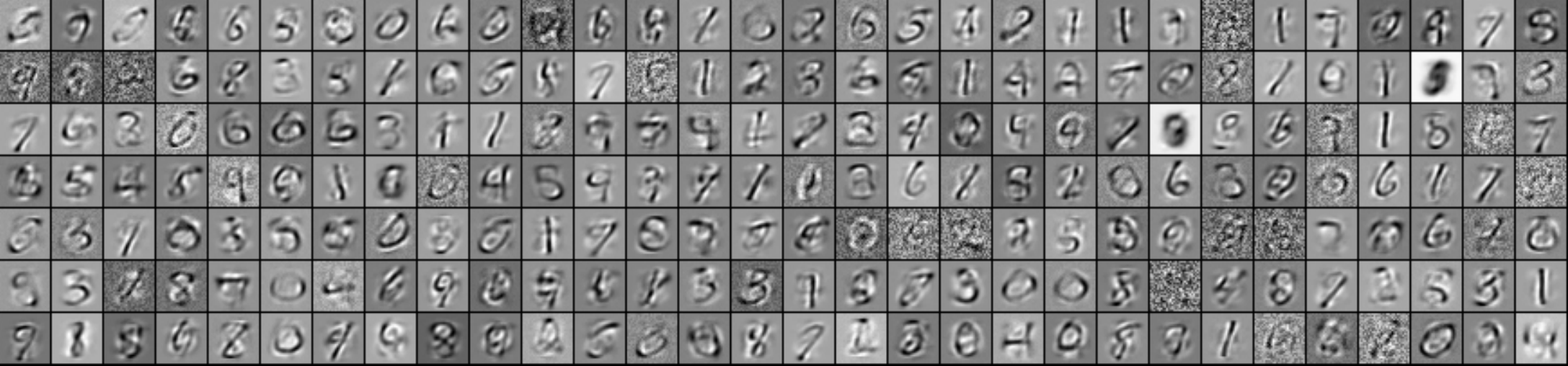}
      \caption{Some entries from the dictionary using the new k-sparse prior with 1000 hidden units and $k=40$}
  \label{fig:k_40_rpca}
\end{figure*}

The learned filters of these two unsupervised models are also very interesting.
Results from the RPCA network using 1000 elements large sparse codes are
shown in \ref{fig:dict_rpca}. Randomly selected entries from the learned
dictionary are shown in this picture. One can see typical filters just like
one would expect it from standard PCA. In \ref{fig:k_20_rpca} and \ref{fig:k_40_rpca}
the dictionaries of two k-sparse coding networks are shown, one with $k=20$ and
the other with $k=40$. In contrast to the RPCA case now one does not see global
filters, but instead, local filters representing line segments of digits. Larger
segments in the $k=20$ case, and smaller ones in the $k=40$ case. These filters
are very similar to those produced by the k-sparse Autoencoder from \cite{makhzani_k-sparse_2013}.

\setlength{\tabcolsep}{2pt}
\begin{table}[ht]

\begin{center}
   \begin{tabular}{c | c | c | c | c}
                &      &k=100 & k=40 & k=20\\ \hline
        RPCA    & 7.80 &      &      &  \\ \hline
        k-sparse &      & 2.87 & 3.27 & 3.45  \\ \hline
    \end{tabular}

\end{center}
\caption{Results for logistic regression on the sparse codes of the efficient RPCA and k-sparse implementation.}
\label{tab:regression}
\end{table}

\section{Conclusion and future work}

The classification quality of an efficient version of RPCA has been presented.
An additional addon was presented solving several problems that arised during the usage
of RPCA. This solution consists of changing the regularizer from a l1-norm to a completely
new prior using the k-sparse function. Due to the mathematical derivation of the network structure
from the objective function this new prior automatically incoperates itself inside the transfer function.
Now the sparse code has a much sparser structure but also the parameter decision
got much more stable. This new k-sparse coding model resulted in much lower classification
errors than the original efficient RPCA version.

Future work consists of testing this new k-sparse norm as prior also for
regular sparse coding or non-negative matrix factorization. Another application
could be to use it as regularizer for any other machine learning algorithm.


\bibliography{literature}
\bibliographystyle{iclr2015}

\end{document}